\def\year{2020}\relax
\documentclass[letterpaper]{article} 
\usepackage{aaai20}  
\usepackage{times}  
\usepackage{helvet} 
\usepackage{courier}  
\usepackage[hyphens]{url}  
\usepackage{graphicx} 
\usepackage{graphicx}  
\usepackage[perpage,multiple]{footmisc}
\usepackage{xcolor}
\usepackage{physics}
\usepackage{subcaption}
\usepackage{multirow}
\usepackage{tabularx}
\usepackage{amsfonts}
\usepackage[absolute]{textpos}

\usepackage{algpseudocode,algorithm,algorithmicx}  
\algrenewcommand\algorithmicrequire{\textbf{Inputs:}}  
\algrenewcommand\algorithmicensure{\textbf{Outputs:}}
\renewcommand{\algorithmiccomment}[1]{\bgroup\hfill//~#1\egroup}

\title{SUOD: Toward Scalable Unsupervised Outlier Detection}
\author{
Yue Zhao,\textsuperscript{\rm 1}
Xueying Ding,\textsuperscript{\rm 1} 
Jianing Yang,\textsuperscript{\rm 2}
Haoping Bai\textsuperscript{\rm 2}\\
\textsuperscript{\rm 1}H. John Heinz III College, Carnegie Mellon University, Pittsburgh, PA 15213, USA\\
\textsuperscript{\rm 2}Machine Learning Department, Carnegie Mellon University, Pittsburgh, PA 15213, USA\\ 
zhaoy@cmu.edu, xding2@andrew.cmu.edu, jianing3@cs.cmu.edu, haopingb@andrew.cmu.edu
}
\begin{document}


\maketitle

\begin{abstract}
Outlier detection is a key field of machine learning for identifying abnormal data objects. Due to the high expense of acquiring ground truth, unsupervised models are often chosen in practice. To compensate for the unstable nature of unsupervised algorithms, practitioners from high-stakes fields like finance, health, and security, prefer to build a large number of models for further combination and analysis. However, this poses scalability challenges in high-dimensional large datasets. In this study, we propose a three-module acceleration framework called \texttt{SUOD} to expedite the training and prediction with a large number of unsupervised detection models. \texttt{SUOD}'s Random Projection module can generate lower subspaces for high-dimensional datasets while reserving their distance relationship. Balanced Parallel Scheduling module can forecast the training and prediction cost of models with high confidence---so the task scheduler could assign nearly equal amount of taskload among workers for efficient parallelization. \texttt{SUOD} also comes with a Pseudo-supervised Approximation module, which can approximate fitted unsupervised models by lower time complexity supervised regressors for fast prediction on unseen data. It may be considered as an unsupervised model knowledge distillation process. Notably, all three modules are independent with great flexibility to ``mix and match"; a combination of modules can be chosen based on use cases. Extensive experiments on more than 30 benchmark datasets have shown the efficacy of \texttt{SUOD}, and a comprehensive future development plan is also presented. 
\end{abstract}

\section{Introduction}

Anomaly detection aims to identify the samples that are deviant from the general data distribution \shortcite{zhao2019pyod}. 
Most of the existing outlier detection algorithms are unsupervised due to the high cost of acquiring ground truth \shortcite{zhao2019lscp}. It is noted that outlier detection can be viewed as a binary classification problem under extreme imbalance (i.e., the number of outliers is way smaller than the number of normal samples). As a result, using a single unsupervised model is risky by nature; it is sensible to use a large group of unsupervised models with variations, e.g., different algorithms, varying parameters, distinct views of the datasets, etc., and more reliable results may be achieved. Outlier ensemble methods that select and combine base detectors are designed for this purpose \shortcite{Aggarwal2013,zimek2014ensembles,Aggarwal2017}. The simplest way is to take the average or maximum values across all the detectors as the final result \shortcite{Aggarwal2017}. More complex combination can also be done in both unsupervised \shortcite{zhao2019lscp} and semi-supervised manners \shortcite{zhao2018xgbod}.

However, both training and predicting with a large number of unsupervised detectors are computationally expensive. This problem is more severe on high-dimensional large samples, especially for proximity-based algorithms that assume outliers behave very differently in specific regions \shortcite{aggarwal2016outlier}. Most algorithms in this category, including \textit{k} nearest neighbors (\textit{k}NN) \shortcite{Ramaswamy2000}, local outlier factor (LOF) \shortcite{Breunig2000}, and local outlier probabilities (LoOP) \shortcite{Kriegel2009}, operate in Euclidean space and suffer from the curse of dimensionality. They can be prohibitively slow or even fail to work completely. The effort has been made to project high-dimensional data to lower subspaces \shortcite{achlioptas2001database}, like simple Principal Component Analysis (PCA) \shortcite{shyu2003novel} and more complex subspace method HiCS \shortcite{keller2012hics}. Engineering cures have been explored as well---the process can be expedited by parallelization on multiple workers (e.g., CPU cores) \shortcite{lozano2005parallel,oku2014parallel}. Recently, knowledge distillation emerges as a popular way of compressing large neural network models \shortcite{hinton2015distilling}, while its usage in outlier detection is still limited.

The aforementioned treatments face various challenges. First, deterministic projection methods, e.g., PCA, are fast but not ideal for building a large number of diversified outlier detectors---it results in the same subspace that cannot induce diversity for outlier ensembles. Complex projection and subspace methods may bring performance improvement for outlier mining, but the generalization capacity is often reduced with strong assumptions. They are not suited for general-purpose outlier detector acceleration.
Existing parallelization learning frameworks can be inefficient if training and prediction task assignments are not balanced among workers. In fact, a group of heterogeneous models can have significantly different computational cost. As a simple example, let us split 100 heterogeneous models into 4 groups (cores) for training. If group \#2 takes significantly longer time than the others to finish, it will behave like the bottleneck of the system. More formally, the imbalanced task scheduling amplifies the impact of slower worker(s) in a system; the system efficiency is curbed by the slowest group. As we have shown in the later section, the existing parallel task scheduling algorithms in popular machine learning libraries like scikit-learn \shortcite{pedregosa2011scikit} may be inefficient under this setting. In addition to these limitations, unsupervised models such as LOF can be slow (high time complexity), or even inappropriate, to predict on unseen data samples by nature. Another downside of unsupervised and non-parametric models is their limited interpretability. In summary, training and predicting with a large number of heterogeneous unsupervised models is computationally expensive, inefficient in parallelization, and limited in interpretability. 

To tap the gap, we propose a three-module acceleration framework called \texttt{SUOD} that leverages random projection, pseudo-supervised approximation, and balanced parallel scheduling for scalability. For high-dimensional data, \texttt{SUOD} generates a random low-dimensional subspace for each unsupervised model by Johnson-Lindenstrauss projection, on which the model is then trained. To improve the training and prediction efficiency in a distributed system, we propose a balanced parallel scheduling heuristic. The key idea is to forecast the running time of each model so that the workload could be evenly distributed among workers. If prediction is needed for unseen data, lower cost supervised regressors are initialized to approximate complex unsupervised models---the supervised models are trained on the original feature space using the outlier scores generated by unsupervised models as the ``pseudo ground truth". The rationale behind is efficient supervised models are faster for prediction and usually more interpretable; it can be viewed as a way of distilling knowledge from unsupervised models \shortcite{hinton2015distilling}. Notably, all these three modules are designed to be fully independent for model acceleration from complementary perspectives. They have great flexibility to be mixed for different needs.

In this work, we make the following contributions:
\begin{enumerate}
    \item Examine the effect of various deterministic and random projection methods on varying size and dimension datasets, and identify the applicable cases of using them for faster execution and diversity induction.
    \item Identify an imbalanced scheduling issue in existing distributed learning systems for heterogeneous detectors and fix it by a new scheduling schema.
    \item Conduct extensive experiments to analyze the results of using pseudo-supervised regression models for approximating unsupervised outlier detectors. To our best knowledge, this is the first research attempt in outlier detection setting.
    \item Demonstrate the effectiveness of the three modules independently, and the extensibility of combining them together as a scalable training and prediction framework.
    \item To foster reproducibility, all code, figure, and datasets used in this study are openly shared\footnote{\url{https://github.com/yzhao062/SUOD}}. A scalable implementation will also be included in PyOD \shortcite{zhao2019pyod} soon. 
    
\end{enumerate}

\section{Related Works}

\subsection{Outlier Detection and Outlier Ensembles}
Anomaly detection has numerous important applications in various fields, such as rare disease detection \shortcite{li2018semi}, fraudulent online review analysis \shortcite{akoglu2013opinion}, and network intrusion detection \shortcite{lazarevic2003comparative}. Despite, detecting outliers is a challenging classification task due to multiple reasons \shortcite{zhao2019lscp}. First, anomalies only consist of a small portion of the entire data--extreme data imbalance incurs difficulty. Second, the limited amount of data and available labels impede learning data representation accurately. Third, the definition of outliers can be ambiguous; outliers may be heterogeneous that should be treated as a multi-class problem.

Most of the existing detection algorithms are unsupervised as ground truth is absent; acquiring labels can be prohibitively expensive in practice. As a result, there are a few established unsupervised anomaly detection algorithms like Isolation Forest \shortcite{liu2008isolation}, Local Outlier Factor (LOF) \shortcite{Breunig2000}, and Angle-based Outlier Detection (ABOD) \shortcite{Kriegel2009}, with different assumptions of the underlying data. Regarding unsupervised deep models like autoencoders and generative adversarial networks \shortcite{liu2019generative}, the amount of accessible data limits their effectiveness on learning representations. No algorithm could always outperform as the assumptions may be incorrect, and it is hard to asses without ground truth.

Therefore, relying on a single unsupervised model has an inherently high risk, and outlier ensembles that leverage a group of detectors become increasingly popular \shortcite{Aggarwal2013,zimek2014ensembles}. There are a group of unsupervised outlier ensemble frameworks proposed in the last several years from simple average, maximization, weighted average, second-phase combination methods \shortcite{Aggarwal2017} to more complex selective models like SELECT \shortcite{rayana2016less} and LSCP \shortcite{zhao2019lscp}. Although unsupervised outlier ensembles methods can be effective in certain cases, they could not incorporate the existing ground truth information regardless of its richness. As a result, a group of semi-supervised detection frameworks that leverage existing labels and enhance the data representation by unsupervised models are proposed. The representative ones include BORE \shortcite{micenkova2014learning} and XGBOD \shortcite{zhao2018xgbod}. They use unsupervised outlier detection scores as additional features to enhance the original feature space, which can be considered as unsupervised feature engineering or representation learning (extraction). It is noted that for both unsupervised and supervised outlier ensembles, a large group diversified unsupervised base detectors are needed---\texttt{SUOD} is therefore designed to facilitate this process.

\subsection{Knowledge Distillation and Model Approximation}
Knowledge distillation refers to the notion of compressing a or an ensemble of large, often cumbersome model(s) into a small and more interpretable model. This is often done by training an ensemble of large models (can be seen as ``Teachers") on a large dataset, followed by using a small and simple model (``Student") to learn the output of the ensemble. There are two main motivations behind knowledge distillation: (i) to reduce the deployment-time computational cost by replacing the large models with a small model and (ii) to increase interpretability as simple models are often more easy to be understood by human. This strategy has seen success in tasks including computer vision \shortcite{romero2014fitnets}, automatic speech recognition \shortcite{hinton2015distilling}, and neural machine translation \shortcite{Kim_2016}. It is noted that the proposed \texttt{SUOD} framework shares a similar concept as knowledge distillation for computational cost optimization but comes with a few fundamental differences as described in Algorithm Design section.

\section{Algorithm Design}

The proposed \texttt{SUOD} contains three independent modules. As shown in Algorithm \ref{alg:SUOD}, the modules may be enabled if specific conditions are met. For high-dimensional data, \texttt{SUOD} randomly project the original input onto lower-dimensional spaces (\textbf{\textit{Module I}}). For expediting the training and prediction with a large number of models, a balanced parallel scheduling mechanism is proposed (\textbf{\textit{Module II}}). If prediction on new samples is needed, an efficient supervised regressor may be initialized to approximate the output of each costly unsupervised detector for prediction (\textbf{\textit{Module III}}).

\subsection{Module I: Random Projection}
A widely used algorithm to alleviate the curse of dimensionality on high-dimensional data is the Johnson-Lindenstraus (JL) projection \shortcite{johnson1984extensions}, although its use in outlier mining is still unexplored. JL projection is a simple compression scheme without heavy distortion on the Euclidean distances of the data. Its built-in randomness is also useful for inducing diversity for outlier ensembles. Despite, projection may be less useful or even detrimental for methods like Isolation Forests and HBOS that rely on subspace splitting.

This linear transformation is defined as: given a set of data $X = \{x_1, x_2, ... x_n\},$ each $x_i \in \mathbb{R}^d$, let $W$ be a $k \times d$ matrix with each entry drawing independently from a $\mathcal{N}(0,1)$ distribution or a Rademacher distribution. Then the JL projection is a function $f: \mathbb{R}^d \rightarrow \mathbb{R}^k$ such that
$f(x_i) = \frac{1}{\sqrt{k}}Wx_i$. JL projection randomly projects high-dimensional data to lower dimension subspaces, but preserve the distance relationship between points. In fact, if we fix some $v \in \mathbb{R}^d$, and let $W$ be the $k \times d$ matrix such that each entry is from $\mathcal{N}(0,1)$. For every 
$\epsilon \in (0,3)$, we have:

\begin{equation}
\label{eq_JL}
P\left[(1-\epsilon){\Vert v \Vert}^2 \leq {\Vert \frac{1}{\sqrt{k}}Wv \Vert}^2 \leq (1 + \epsilon) {\Vert v \Vert}^2 \right] \leq 2e^{-\epsilon^2 \frac{k}{6}}
\end{equation}

Furthermore, fix $v$ to be the differences between vectors. Then, the above bound also shows that for a finite set of $N$ vectors $X = \{x_1, x_2, ... x_n\} \subseteq \mathbb{R}^d$, the pairwise Euclidean distance is preserved within a factor of $(1 \pm \epsilon)$, if we reduce the vectors to $k = \mathcal{O}(\frac{log(N)}{\epsilon ^2})$ dimensions.

Four JL projection methods are therefore introduced for their great property in compression and diversity induction: (i) \textit{basic}: the transformation matrix is generated by standard Gaussian; (ii) \textit{discrete}: the transformation matrix is picked  randomly from Rademacher distribution (uniform in $\{-1,1\})$; (iii) \textit{circulant}: the transformation matrix is obtained by rotating the subsequent rows from the first row which is generated from standard Gaussian and (iv) \textit{toeplitz}: the first row and column of the transformation matrix are generated from standard Gaussian, and each diagonal uses a constant value from the first row and column. 

Let $X \in \mathcal{R}^{n\times d}$ denote a dataset with $n$ points and $d$ features. In this work, we only invoke the projection if the data dimension exceeds 20 (dimension threshold $\theta = 20$) and reduce the original dimension by half (projection dimension $k=\frac{1}{2}d$). \texttt{SUOD} check whether $d$ is higher the projection threshold $\theta$. If so, a JL transformation matrix $W \in \mathcal{R}^{k\times d}$ is initialized by one of the JL projection methods and $X$ is projected onto the $k$ dimension subspace by $W$.



\subsection{Module II: Balanced Parallel Scheduling}
\label{balancedLearning}

Balanced Parallel Scheduling (BPS) aims for assigning training and prediction tasks more evenly based on the model costs, across all available workers. For instance, one may train 25 detectors with varying parameters from each of the four algorithm groups \{\textit{k}NN, LOF, ABOD, OCSVM\}, resulting in 100 models in total. The existing parallel frameworks, e.g., the voting machine in scikit-learn \shortcite{pedregosa2011scikit}, will simply split the the models into 4 subgroups by order and schedule the first 25 models (all \textit{k}NNs) on Core 1 (worker 1), the next 25 models on Core 2, etc. This does not account for the fact that within a group of heterogeneous detectors, the computational cost varies. Scheduling the task with equal number of models can result in highly imbalanced tasks. In the worst case scenario, one worker may be significantly slower than the rest, and the entire process halts. Obviously, this problem applies to both training and prediction stage.

\begin{figure}[t]
\caption{\label{fig:flowchart}Flowchart of Balanced Parallel Scheduling} 
\label{fig:balanceLearning}
\centering
    \includegraphics[width=\linewidth]{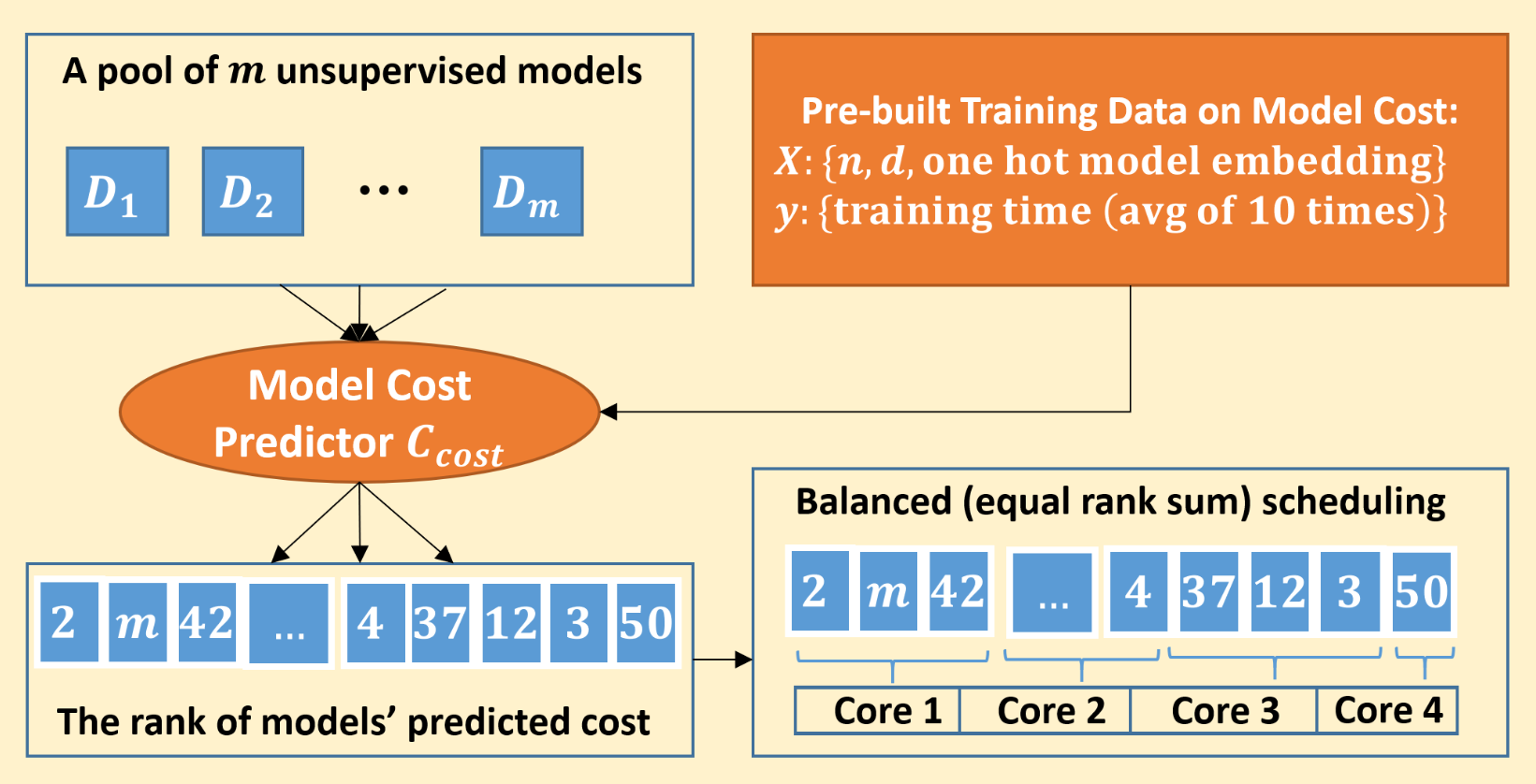}
\end{figure}

The proposed BPS heuristic focuses on designing a more balanced schedule among workers. Ideally, all workers can finish the scheduled tasks within the similar duration of time and return the result. As shown in Fig. \ref{fig:balanceLearning}, we build a \textit{model cost predictor} $C_{cost}$ to forecast the model running time (sum of 10 trails) given the input data size, input data dimension, and the algorithm embedding (one-hot). Given the time and resource limitation, we built a training set with 11 algorithms on 47 benchmark datasets (see Experiment section for details), and a random forest regressor \shortcite{breiman2001random} is trained on the dataset with 10-fold cross validation. Although the fitted regressor does not have a high $R^2$ score, the Spearman's Rank correlation \shortcite{spearman1904proof} consistently shows high value ($r_s>0.9$) with low p-value ($p<0.0001$). This implies that even the cost predictor $C_{cost}$ could not predict the running time precisely, it can predict the rank of the running time with high accuracy. As a result, we propose a scheduling heuristic by enforcing nearly equal sum of the rank on running time. Given there are $m$ models to be trained, cost predictor $C_{cost}$ is first invoked to forecast the time cost for a given model $\mathcal{D}_i$ in $\mathcal{D}$ as $C_{cost}(\mathcal{D}_i)$. After the prediction is done, the predicted time is converted to a rank in $[1,m]$. If there are $t$ cores (workers), each worker will be assigned a group of models to achieve the objective of minimizing the workload imbalance among workers (Eq. \ref{eq:balanced_learning_goal}). So each group has a rank sum at around $\frac{m^2+m}{2t}$. One additional advantage of using rank is transferability: the running time will vary on different machines but the relative rank should preserve.



\begin{equation}
\label{eq:balanced_learning_goal}
    \min_{\mathcal{W}} \sum_{i=1}^{t} \abs{\sum_{D_j \in \mathcal{W}_i}{C_{cost}(\mathcal{D}_i)-\sum_{l=1}^{m}C_{cost}(\mathcal{D}_l)}}
\end{equation}

\subsection{Module III: Pseudo-Supervised Approximation}
\label{al:pseudo_approximation}
Once the unsupervised models have been fitted on the reduced feature space generated in \textbf{\textit{Module I}} or the original space (if no random projection is involved). \texttt{SUOD} can approximate and replace each of \textbf{costly unsupervised model} by a \textbf{faster supervised regression model} for predicting on unseen samples. In other words, not all unsupervised models should be replaced but only the expensive ones. There is no efficiency incentive to approximate fast algorithms like Isolation Forest. However, most of the proximity-based algorithms like \textit{k}NN and LOF have high time complexity for prediction (upper bounded by $\Theta (nd)$), which can be effectively replaced by fast supervised models like random forest (upper bounded by $\Theta (dp)$ where $p$ denotes the number of base trees). This ``pseudo-supervised" model uses the output of unsupervised models as ``the pseudo ground truth"---the goal is to approximate and find a better mapping from the original input to ``the output of an unsupervised model". Ensemble-based tree models are recommended for their outstanding scalability, robustness to overfitting, and interpretability (e.g., feature importance). Notably, this process can also be viewed as using supervised regressors to distill knowledge from unsupervised models. However, our approximation is different from the established knowledge distillation mainly in three aspects. First, our approximation works in a fully unsupervised manner unlike the classic distillation under supervised settings. Second, our ``teacher" and ``student" models have totally different architectures with little to no similarity. For instance, we use random forest (an ensemble tree model) to approximate LOF (a proximity-based density estimation model). Third, our approximation leads to a clear interpretability improvement, whereas the student models in neural networks lack that. See Appendix \ref{append:approx_advantage}.

As shown in Algorithm \ref{alg:SUOD}, for each trained unsupervised model $\mathcal{D}_i$, a supervised regressor $\mathcal{R}_i$ might be built with the original input $X$ and the pseudo ground truth (the output of $\mathcal{D}_i$). The prediction on unseen data will be then made by $\mathcal{R}$. This approximation shows multiple benefits:

\begin{enumerate}
    \item Compared with non-parametric unsupervised models, parametric supervised models may have lower space complexity and faster prediction speed.
    \item Supervised models generally show better interpretability compared with unsupervised counterparts. For instance, random forest used in this work can generate feature importance automatically to facilitate understanding.
    \item Not all unsupervised models are appropriate for making prediction. Taking LOF as an example, the fitted model is not supposed to predict on unseen data. A supervised approximator may be used for prediction in this case.
\end{enumerate}

\begin{algorithm} [!ht]
	\caption{Scalable Unsupervised Outlier Detection}
	\label{alg:SUOD}
	\begin{algorithmic}[1]
		\Require{a group of $m$ unsupervised outlier detectors $\mathcal{D}$; $d$ dimension training data $X_{train}$ and test data $X_{test}$ (optional); projection threshold $\theta$; projection dimension $k$; pre-trained cost predictor $C_{cost}$; \# of workers $t$}
		\Ensure{fitted unsupervised models $\mathcal{D}$; fitted pseudo-supervised regressors $\mathcal{R}$ (optional); prediction results on test data $\hat{y}_{test}$ (optional)}
		\For{Each detector $\mathcal{D}_i$ in $\mathcal{D}$}
		\If { $d > \theta$} \Comment{\textit{enable random projection}}
		\State Generate random subspace $\psi_{i}$ by JL projection on $X_{train}$ (see \textbf{\textit{Module I}})
		\Else{ $d <\theta$} \Comment{\textit{disable random projection}}
		\State Use the original feature space $\psi_{i} \mathrel{\mathop:}=X_{train}$ 
		\EndIf
		\EndFor
		\If {Parallel learning == True}
		\State Train each $\mathcal{D}_i$ on the corresponding $\psi_{i}$ by Balanced Parallel Scheduling on $t$ workers (Eq. (\ref{eq:balanced_learning_goal}) and Fig. 1)
		\EndIf
		\State Return $\mathcal{D}$
		\If {$X_{test}$ is presented \textbf{and} Approximation == True}
		\State Acquire the pseudo ground truth $target^{\psi_{i}}$ as the output of $\mathcal{D}_i$ on $\psi_{i}$: $target^{\psi_{i}} \mathrel{\mathop:}=\mathcal{D}_i(\psi_{i})$ 
		\For{Each detector $\mathcal{D}_i$ in $\mathcal{D}$}
        \State Initialize a supervised regressor $\mathcal{R}_i$
		\State Fit $\mathcal{R}_i$ by $\{X, target^{\psi_{i}}\}$ in pseudo-supervised approximation described in \textbf{\textit{Module III}}
		\State $\hat{y}_{test}^{i}=\mathcal{R}_i.\textrm{predict}(X_{test})$
		\EndFor
		\State Return $\hat{y}_{test}$ and fitted regressors $\mathcal{R}$
		\EndIf
	\end{algorithmic}
\end{algorithm}

\section{Numerical Experiments and Discussion}

In this preliminary study, three independent experiments are conducted to understand: (i) how will random projection methods affect the performance of outlier detection algorithms; (ii) whether the proposed parallel scheduling algorithm brings performance improvement over the existing approaches and (iii) will pseudo-supervised models lead to degraded prediction performance compared with the original unsupervised models? Because all three modules are independent and can be combined seamlessly, it is assumed that the (partly) combined framework should also work if individual components manage to work. 

\subsection{Datasets, Evaluation Metrics, and Implementation}
More than 30 outlier detection benchmark datasets are used in this study\footnote{ODDS Library: \url{http://odds.cs.stonybrook.edu}}\footnote{DAMI Datasets: \url{http://www.dbs.ifi.lmu.de/research/outlier-evaluation/DAMI}}); the detail is available on online supplementary due to the space limit. The data size $n$ varies from 219 (\textbf{Glass}) to 567,479 (\textbf{HTTP}) samples and the dimension $d$ ranges from 3 to 400.
For both random projection and parallel scheduling experiments, full datasets are used for training. For the pseudo-supervised approximation experiments, 60\% of the data is used for training and the remaining 40\% is set aside for validation. For all experiments, performance is evaluated by taking the average of 10 independent trials using area under the receiver operating characteristic (ROC) and precision at rank n (P@N). Both metrics are widely used in outlier research \shortcite{liu2008isolation,zimek2014ensembles,rayana2016less,Aggarwal2017,zhao2018xgbod,zhao2019lscp,liu2019generative}.

All the unsupervised models are from Python Outlier Detection Toolbox (\texttt{PyOD}), a popular library for outlier mining \shortcite{zhao2019pyod}. Supervised regressors and utility functions are from standard libraries (\texttt{scikit-learn} and \texttt{numpy}). For a fair comparison, none of the models involve parameter tuning process---the default values are used. As all three modules involve time profiling, the same machine (Intel i7-9700 @ 3.00 GHZ; 32 GB RAM) is used for a fair comparison.

\begin{table*}[htb]
    \caption{Comparison of various projection methods on different outlier detectors and datasets}
    \label{table:projection_comparison}
    \begin{subtable}[t]{.33\textwidth}
        \centering
        \caption{ABOD on \textbf{MNIST}}
            \begin{tabular}{@{\extracolsep{1pt}}lcccc}
                \hline
                \textbf{Method}	&	\textbf{Time}	&	\textbf{ROC}		&   \textbf{PRN}     \\
                \hline 
                original	&	12.89	&	0.80	&	0.39	\\
                PCA	        &	8.93	&	0.81	&	0.37	\\
                RS	        &	8.27	&	0.74	&	0.32	\\
                \textit{basic}	&	8.94	&	0.80	&	0.38	\\
                \textit{discrete}	&	8.86	&	0.80	&	0.39	\\
                \textit{circulant}	&	9.33	&	0.80	&	0.38	\\
                \textit{toeplitz}	&	8.96	&	0.80	&	0.38	\\
                \hline \\
            \end{tabular}
    \end{subtable}%
   \begin{subtable}[t]{.33\textwidth}
        \centering
        \caption{ABOD on \textbf{Satellite}}
            \begin{tabular}{@{\extracolsep{1pt}}lcccc}
                \hline
                \textbf{Method}	&	\textbf{Time}	&	\textbf{ROC}		&   \textbf{PRN}     \\
                \hline 
                original	&	4.03	&	0.59	&	0.41	\\
                PCA	&	3.01	&	0.62	&	0.44	\\
                RS &	3.53	&	0.63	&	0.44	\\
                \textit{basic}	&	3.10	&	0.64	&	0.45	\\
                \textit{discrete}	&	3.12	&	0.65	&	0.46	\\
                \textit{circulant}	&	3.14	&	0.66	&	0.48	\\
                \textit{toeplitz}	&	3.14	&	0.66	&	0.47	\\
        \hline \\
        \end{tabular}
    \end{subtable}
   \begin{subtable}[t]{.33\textwidth}
        \centering
        \caption{ABOD on \textbf{Satimage-2}}
            \begin{tabular}{@{\extracolsep{1pt}}lcccc}
                \hline
                \textbf{Method}	&	\textbf{Time}	&	\textbf{ROC}		&   \textbf{PRN}     \\
                \hline 
                original	&	3.68	&	0.85	&	0.28	\\
                PCA	&	2.70	&	0.88	&	0.30	\\
                RS &	3.20	&	0.89	&	0.28	\\
                \textit{basic}	&	2.78	&	0.91	&	0.29	\\
                \textit{discrete}	&	2.79	&	0.91	&	0.31	\\
                \textit{circulant}	&	2.85	&	0.91	&	0.29	\\
                \textit{toeplitz}	&	2.83	&	0.92	&	0.30	\\
        \hline \\
        \end{tabular}
    \end{subtable}
    \begin{subtable}[t]{.33\textwidth}
        \centering
        \caption{LOF on \textbf{MNIST}}
            \begin{tabular}{@{\extracolsep{1pt}}lcccc}
                \hline
                \textbf{Method}	&	\textbf{Time}	&	\textbf{ROC}		&   \textbf{PRN}     \\
                \hline 
                original	&	7.64	&	0.68	&	0.29	\\
                PCA	&	4.92	&	0.67	&	0.27	\\
                RS &	3.65	&	0.63	&	0.23	\\
                \textit{basic}	&	4.87	&	0.70	&	0.31	\\
                \textit{discrete}	&	5.21	&	0.70	&	0.32	\\
                \textit{circulant}	&	5.06	&	0.69	&	0.31	\\
                \textit{toeplitz}	&	4.97	&	0.71	&	0.31	\\
                \hline \\
            \end{tabular}
    \end{subtable}%
   \begin{subtable}[t]{.33\textwidth}
        \centering
        \caption{LOF on \textbf{Satellite}}
            \begin{tabular}{@{\extracolsep{1pt}}lcccc}
                \hline
                \textbf{Method}	&	\textbf{Time}	&	\textbf{ROC}		&   \textbf{PRN}     \\
                \hline 
                original	&	0.82	&	0.55	&	0.38	\\
                PCA	&	0.23	&	0.54	&	0.36	\\
                RS &	0.39	&	0.54	&	0.37	\\
                \textit{basic}	&	0.31	&	0.54	&	0.37	\\
                \textit{discrete}	&	0.32	&	0.54	&	0.37	\\
                \textit{circulant}	&	0.39	&	0.55	&	0.37	\\
                \textit{toeplitz}	&	0.37	&	0.54	&	0.37	\\
        \hline \\
        \end{tabular}
    \end{subtable}
   \begin{subtable}[t]{.33\textwidth}
        \centering
        \caption{LOF on \textbf{Satimage-2}}
            \begin{tabular}{@{\extracolsep{1pt}}lcccc}
                \hline
                \textbf{Method}	&	\textbf{Time}	&	\textbf{ROC}		&   \textbf{PRN}     \\
                \hline 
                original	&	0.79	&	0.54	&	0.07	\\
                PCA	&	0.20	&	0.52	&	0.04	\\
                RS &	0.37	&	0.53	&	0.08	\\
                \textit{basic}	&	0.29	&	0.52	&	0.08	\\
                \textit{discrete}	&	0.30	&	0.53	&	0.07	\\
                \textit{circulant}	&	0.43	&	0.59	&	0.11	\\
                \textit{toeplitz}	&	0.32	&	0.54	&	0.09	\\
        \hline \\
        \end{tabular}
    \end{subtable}
 \begin{subtable}[t]{.33\textwidth}
        \centering
        \caption{\textit{k}NN on \textbf{MNIST}}
            \begin{tabular}{@{\extracolsep{1pt}}lcccc}
                \hline
                \textbf{Method}	&	\textbf{Time}	&	\textbf{ROC}		&   \textbf{PRN}     \\
                \hline 
                original	&	7.13	&	0.84	&	0.42	\\
                PCA	&	3.92	&	0.84	&	0.40	\\
                RS &	3.33	&	0.77	&	0.34	\\
                \textit{basic}	&	4.17	&	0.84	&	0.42	\\
                \textit{discrete}	&	4.11	&	0.84	&	0.41	\\
                \textit{circulant}	&	4.13	&	0.84	&	0.41	\\
                \textit{toeplitz}	&	4.11	&	0.84	&	0.42	\\
                \hline \\
            \end{tabular}
    \end{subtable}%
   \begin{subtable}[t]{.33\textwidth}
        \centering
        \caption{\textit{k}NN on \textbf{Satellite}}
            \begin{tabular}{@{\extracolsep{1pt}}lcccc}
                \hline
                \textbf{Method}	&	\textbf{Time}	&	\textbf{ROC}		&   \textbf{PRN}     \\
                \hline 
                original	&	0.71	&	0.67	&	0.49	\\
                PCA	&	0.18	&	0.67	&	0.50	\\
                RS &	0.31	&	0.68	&	0.49	\\
                \textit{basic}	&	0.24	&	0.68	&	0.49	\\
                \textit{discrete}	&	0.25	&	0.69	&	0.50	\\
                \textit{circulant}	&	0.33	&	0.70	&	0.50	\\
                \textit{toeplitz}	&	0.30	&	0.70	&	0.51	\\
        \hline \\
        \end{tabular}
    \end{subtable}
   \begin{subtable}[t]{.33\textwidth}
        \centering
        \caption{\textit{k}NN on \textbf{Satimage-2}}
            \begin{tabular}{@{\extracolsep{1pt}}lcccc}
                \hline
                \textbf{Method}	&	\textbf{Time}	&	\textbf{ROC}		&   \textbf{PRN}     \\
                \hline 
                original	&	0.68	&	0.94	&	0.39	\\
                PCA	&	0.15	&	0.94	&	0.39	\\
                RS &	0.29	&	0.94	&	0.38	\\
                \textit{basic}	&	0.23	&	0.94	&	0.38	\\
                \textit{discrete}	&	0.20	&	0.95	&	0.37	\\
                \textit{circulant}	&	0.36	&	0.96	&	0.37	\\
                \textit{toeplitz}	&	0.25	&	0.96	&	0.39	\\
        \hline \\
        \end{tabular}
    \end{subtable}
\end{table*}

\subsection{Comparison among Projection Methods}
To evaluate the effect of projection methods on outlier detector performance, we choose three expensive detection algorithms (ABOD, LOF, and \textit{k}NN) to measure their execution time, ROC, and P@N with different projection methods. All three methods directly or indirectly measure sample similarity in the Euclidean space , e.g., pairwise sample distances, which is susceptible to the curse of dimensionality and projection may be particularly helpful. 

Table \ref{table:projection_comparison} shows the comparison among four JL projection variations with original (no projection is used), PCA, and RS (randomly select $k$ features from the original $d$ features, used in Feature Bagging \shortcite{lazarevic2005feature} and LSCP). First, all projection methods show superiority regarding time cost. Second, using RS method comes with high instability, and shows performance decrease on all three datasets. This observation agrees with the finding by Zhao et al. \shortcite{zhao2019lscp}, in which they used an ensemble projection to overcome the instability. Third, PCA is slightly faster than JL projection methods, although the detector performance by PCA projection is not as good as the ones with JL projections as shown in subtable (b)-(i). Moreover, PCA as a deterministic method, may not be ideal for inducing diversity in outlier ensembles, as it always result in the same sets of subspaces. Fourth, among all four JL projection methods, \textit{circulant} and \textit{toeplitz} outperform in most cases. Since \textit{toeplitz} is slightly faster than \textit{circulant}, it is a reasonable choice for reducing dimensionality and inducing diversity for the models that are susceptible to the curse of dimensionality.

\subsection{The Effect of Balanced Parallel Scheduling}

To verify the effectiveness of the proposed BPS algorithm, we run the following experiments by varying: (i) the size ($n$) and the dimension ($d$) of the datasets, (ii) the number of estimators to train ($m$) and (iii) the number of CPU cores ($t$). The time elapsed is measured in seconds. Due to the space limit, we only show the comparison between the simple scheduling and BPS on \textbf{Cardio} ($n=1831, d=21$), \textbf{PageBlocks} ($n=5393, d=10$), and \textbf{Pendigits} ($n=6870, d=16$), by setting $t \in \{2,4,6\}$ and $m \in \{100,500\}$. 

More results can be found on the online supplementary, and the conclusion holds for all tested datasets. Table \ref{table:BPS} shows that the proposed BPS has a clear edge over the simple scheduling mechanism (denoted as \textbf{Simple} in the tables) that equally splits the tasks by order. It yields a significant time reduction (denoted as \textbf{\% RED} in the tables), and gets more significant if more estimators and cores are involved. For instance, if 500 estimators and 6 cores are used, the time reduction over the simple scheduling is more than 40\%. This agrees with our assumption that the imbalanced task scheduling will lead to more inefficient consequences with the increasing number of estimators and workers, and will therefore benefit more from the proposed BPS heuristic. 

\subsection{The Analysis of Pseudo-Supervised Approximation}
\begin{figure*}[t]
\caption{Comparison among unsupervised models and their pseudo-supervised counterparts} 
\label{fig_all_comparision}
\centering
    \includegraphics[width=\linewidth]{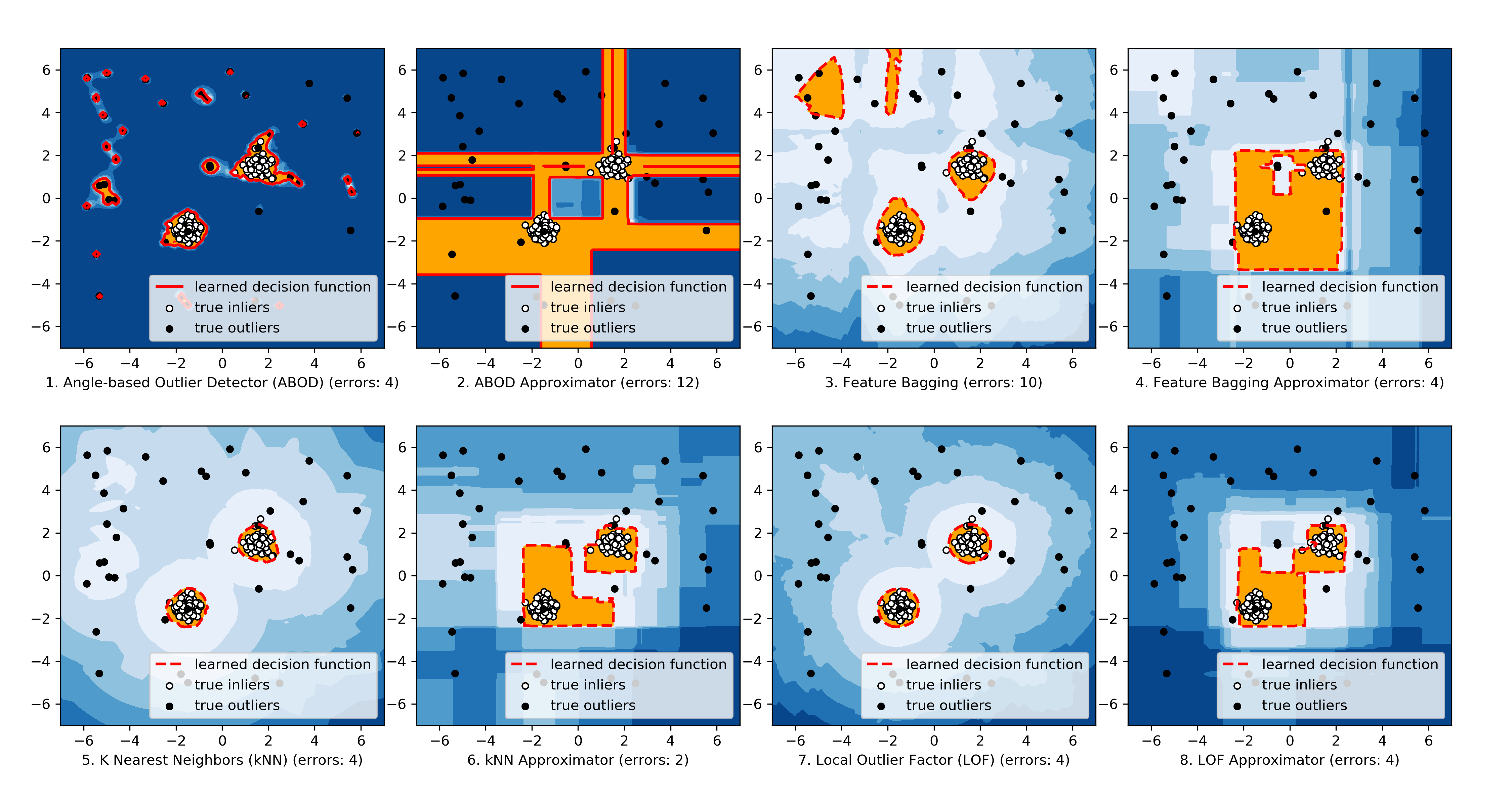}
\end{figure*}

To better understand the behavior of the pseudo-supervised approximation, we first generate 200 synthetic points with Normal distribution for outliers (40 points) and Uniform distribution for normal samples (160 points). In Fig. \ref{fig_all_comparision}, we plot the decision surfaces of unsupervised models and their supervised approximators (random forest regressor). It is clear that the decision surfaces are different and some regularization effect appears (lower errors on Feature Bagging and \textit{k}NN). One of the assumptions is that the approximation process improves the generalization ability of the model by ``ignoring" the overfitted points. This fails to work with ABOD because it has a extremely coarse decision surfaces to approximate (see Fig. \ref{fig_all_comparision}).

Table \ref{table:approx_roc} and \ref{table:approx_p@n} compare prediction performance between the original unsupervised models and pseudo-supervised approximators on 8 datasets with 6 algorithms. These algorithms are known to be more computationally expensive than the supervised random forest regressors. The approximators with performance degradation are highlighted in bold and italicized in the tables. The prediction time comparison is omitted due to space limit, but the gain is clear (see online supplementary). Therefore, the focus is put on prediction ROC and P@N to see whether the approximators could predict unseen data as good as the original unsupervised models. The acceptable threshold of performance degradation between an approximator and its original unsupervised models is set as $[0,0.01]$ and any negative difference larger than $0.01$ will be regarded as degradation. The tables reveal that not all the algorithms can be approximated well by supervised regressors: ABOD has a performance decrease regarding ROC on multiple datasets. ABOD is a linear models that look for a lower-dimensional subspace to embed the normal samples \shortcite{aggarwal2016outlier}, so the approximation may not work if it has extremely complex decision surfaces as mentioned before. In contrast, proximity-based models that aim to identify specific Euclidean regions in which outliers are different, benefit from the approximation. Both table shows, \textit{k}NN, LoF, and A\textit{k}NN (average \textit{k}NN) experience an performance gain. Specifically, all three algorithms yield around 100\% ROC increase on \textbf{HTTP}. Other algorithms, such as Feature Bagging and CBLOF, the ROC and PRC performances stay within the acceptable range. In other words, it is useful to perform pseudo-supervised approximation for these estimators as the time efficiency is improved at little to no loss in accuracy. Through the visualization and quantitative comparisons, we believe that the proposed pseudo-supervised approximation is meaningful for prediction acceleration.

\begin{table}
\centering
	\caption{Comparison between simple scheduling and BPS} 
	\begin{tabular}{c  c | c  c |  r  r r } 
		\hline 
		$n$ & $d$ & $m$ & $t$ & \textbf{Simple} & \textbf{BPS} & \textbf{\% RED}\\
		\hline
        1831	&	21	&	100	&	2	&	26.33	&	19.85	&	24.61 \\
        1831	&	21	&	100	&	4	&	17.93	&	13.69	&	23.65 \\
        1831	&	21	&	100	&	6	&	19.16	&	15.23	&	20.51 \\
        1831	&	21	&	500	&	2	&	100.51	&	72.16	&	28.21 \\
        1831	&	21	&	500	&	4	&	80.38	&	39.46	&	50.91 \\
        1831	&	21	&	500	&	6	&	55.3	&	32.78	&	40.72 \\
        5393	&	10	&	100	&	2	&	51.11	&	35.17	&	31.19 \\
        5393	&	10	&	100	&	4	&	42.49	&	16.23	&	61.80 \\
        5393	&	10	&	100	&	6	&	38.45	&	16.97	&	55.86 \\
        5393	&	10	&	500	&	2	&	197.84	&	137.46	&	30.52 \\
        5393	&	10	&	500	&	4	&	167.36	&	76.14	&	54.51 \\
        5393	&	10	&	500	&	6	&	127.08	&	66.29	&	47.84 \\
        6870	&	16	&	100	&	2	&	80.89	&	67.23	&	16.89 \\
        6870	&	16	&	100	&	4	&	68.31	&	35.29	&	48.34 \\
        6870	&	16	&	100	&	6	&	49.19	&	24.18	&	50.84 \\
        6870	&	16	&	500	&	2	&	336.54	&	286.14	&	14.98 \\
        6870	&	16	&	500	&	4	&	345.69	&	164.02	&	52.55 \\
        6870	&	16	&	500	&	6	&	211.34	&	113.13	&	46.47 \\

		\hline
	\end{tabular}
	\label{table:BPS} 
\end{table}

\section{Conclusion and Future Directions}

In this work, a three-module framework called \texttt{SUOD} is proposed to accelerate the training and prediction with a large number of unsupervised anomaly detectors. The three modules in \texttt{SUOD} focus on different perspectives of scalability enhancement: (i) Random Projection module generates lower-dimensional subspaces to alleviate the curse of dimensionality using toeplitz Johnson-Lindenstraus projection; (ii) Balanced Parallel Scheduling module ensures that nearly equal amount of workloads are assigned to multiple workers in parallel training and prediction and (iii) Pseudo-supervised Approximation module could accelerate costly unsupervised models' prediction speed by replacing them by scalable supervised regressors, which also brings the extra benefit regarding interpretability and storage cost. The extensive experiments on more than 30 benchmark datasets empirically show the great potential of \texttt{SUOD}, and many intriguing results are observed. To improve the model reproducibility and accessibility, all code, figures, implementation for demo and production, will be released\footnote{\url{https://github.com/yzhao062/SUOD}}.

Many investigations are underway. First, we would like to demonstrate \texttt{SUOD}'s effectiveness as an end-to-end framework by combining three modules, and provide an easy to use toolkit in Python. Three additional experiments may be conducted to show that \texttt{SUOD} is useful for: (i) simple combination like majority vote and maximization \shortcite{zhao2020combo}; (ii) more complex unsupervised model combination like LSCP \shortcite{zhao2019lscp} and (iii) supervised outlier combination algorithms such as XGBOD \shortcite{zhao2018xgbod}. Second, although we provide a pre-built \textit{model cost predictor} as part of the framework, a re-trainable cost predictor is expected so practitioners can make accurate prediction on their machines. Third, we would further emphasize the interpretability provided by the pseudo-supervised approximation, which can be beyond simple feature importance provided in tree regressors. Fourth, we see there is room to investigate why and how the pseudo-supervised approximation could work in a more strict and theoretical way. Specifically, we want to know how to choose supervised regressors and under what conditions the approximation could work. This study, as the first step, empirically shows that proximity-based models benefit from the approximation, whereas linear models may not.
Lastly, we want to verify that \texttt{SUOD} can work with real-world use cases. We are in contact with a pharmaceutical consultancy firm to access their rare disease detection datasets (which can be viewed as an outlier detection task). Microsoft Malware Dataset\footnote{\url{https://www.kaggle.com/c/microsoft-malware-prediction}} is also chosen to explore \texttt{SUOD}'s applicability.

\begin{table*}[ht]\centering
	\caption{Test ROC scores of unsupervised models and their pseudo-supervised approximators (avg of 10 independent trials)} 
	\label{table:approx_roc}
	\footnotesize
	\begin{tabular}{l c c c c c c c c c c c c c c c c} 
		\hline 
		\textbf{Dataset} & \multicolumn{2}{c}{\textbf{Annthyroid}} & \multicolumn{2}{c}{\textbf{Breastw}} & \multicolumn{2}{c}{\textbf{Cardio}} & \multicolumn{2}{c}{\textbf{HTTP}} &  \multicolumn{2}{c}{\textbf{MNIST}} & \multicolumn{2}{c}{\textbf{Pendigits}} & \multicolumn{2}{c}{\textbf{Pima}} & \multicolumn{2}{c}{\textbf{Satellite}} \\
		\hline
		\textbf{Model} & Orig & Appr & Orig & Appr & Orig & Appr & Orig & Appr & Orig & Appr & Orig & Appr & Orig & Appr & Orig & Appr\\
        ABOD	&	0.83	&	\textbf{\textit{0.71}}	&	0.92	&	0.93	&	0.63	&	\textbf{\textit{0.53}}	&	0.15	&	\textbf{\textit{0.13}}	&	0.81	&	\textbf{\textit{0.79}}	&	0.67	&	0.82	&	0.66	&	0.70	&	0.59	&	0.68	\\
        CBLOF	&	0.67	&	0.68	&	0.96	&	0.98	&	0.73	&	0.76	&	1.00	&	1.00	&	0.85	&	0.89	&	0.93	&	0.93	&	0.63	&	0.68	&	0.72	&	0.77	\\
        FB	&	0.81	&	\textbf{\textit{0.45}}	&	0.34	&	\textbf{\textit{0.10}}	&	0.61	&	0.70	&	0.34	&	0.97	&	0.72	&	0.83	&	0.39	&	0.51	&	0.59	&	0.63	&	0.53	&	0.64	\\
        KNN	&	0.80	&	0.79	&	0.97	&	0.97	&	0.73	&	0.75	&	0.19	&	0.85	&	0.85	&	0.86	&	0.74	&	0.87	&	0.69	&	0.71	&	0.68	&	0.75	\\
        AKNN	&	0.81	&	0.82	&	0.97	&	0.97	&	0.67	&	0.72	&	0.19	&	0.88	&	0.84	&	0.85	&	0.72	&	0.87	&	0.69	&	0.71	&	0.66	&	0.74	\\
        LOF	&	0.74	&	0.85	&	0.44	&	0.45	&	0.60	&	0.68	&	0.35	&	0.75	&	0.72	&	0.76	&	0.38	&	0.47	&	0.59	&	0.65	&	0.53	&	0.66	\\
    \hline 
	\end{tabular}
\end{table*}

\begin{table*}[ht]\centering
	\caption{Test P@N scores of unsupervised models and their pseudo-supervised approximators (avg of 10 independent trials)} 
	\label{table:approx_p@n}
	\footnotesize
	\begin{tabular}{l c c c c c c c c c c c c c c c c} 
		\hline 
		\textbf{Dataset} & \multicolumn{2}{c}{\textbf{Annthyroid}} & \multicolumn{2}{c}{\textbf{Breastw}} & \multicolumn{2}{c}{\textbf{Cardio}} & \multicolumn{2}{c}{\textbf{HTTP}} &  \multicolumn{2}{c}{\textbf{MNIST}} & \multicolumn{2}{c}{\textbf{Pendigits}} & \multicolumn{2}{c}{\textbf{Pima}} & \multicolumn{2}{c}{\textbf{Satellite}} \\
		\hline
		\textbf{Model} & Orig & Appr & Orig & Appr & Orig & Appr & Orig & Appr & Orig & Appr & Orig & Appr & Orig & Appr & Orig & Appr\\
        ABOD	&	0.31	&	\textbf{\textit{0.08}}	&	0.80	&	0.83	&	0.27	&	\textbf{\textit{0.20}}	&	0.00	&	0.00	&	0.40	&	0.27	&	0.05	&	0.05	&	0.48	&	0.52	&	0.41	&	0.46	\\
        CBLOF	&	0.25	&	\textbf{0.24}	&	0.86	&	0.90	&	0.31	&	0.34	&	0.02	&	\textbf{0.01}	&	0.42	&	0.48	&	0.35	&	0.36	&	0.43	&	0.48	&	0.54	&	0.57	\\
        FB	&	0.24	&	\textbf{\textit{0.02}}	&	0.03	&	0.07	&	0.23	&	0.26	&	0.02	&	0.04	&	0.34	&	0.36	&	0.03	&	0.07	&	0.37	&	0.44	&	0.37	&	0.42	\\
        KNN	&	0.30	&	0.32	&	0.89	&	0.89	&	0.37	&	0.46	&	0.03	&	0.03	&	0.42	&	0.45	&	0.08	&	\textbf{\textit{0.06}}	&	0.47	&	0.47	&	0.49	&	0.53	\\
        AKNN	&	0.30	&	0.33	&	0.88	&	0.89	&	0.34	&	0.40	&	0.03	&	0.03	&	0.41	&	0.45	&	0.05	&	0.13	&	0.48	&	0.49	&	0.47	&	0.52	\\
        LOF	&	0.27	&	0.36	&	0.19	&	0.35	&	0.23	&	0.23	&	0.01	&	0.03	&	0.33	&	0.32	&	0.03	&	0.08	&	0.40	&	0.44	&	0.37	&	0.42	\\
    \hline 
	\end{tabular}
\end{table*}

\medskip

%
%
\bibliographystyle{aaai}
\bibliography{bibfile}

\end{document}